\documentclass{svproc}

\usepackage{graphicx}
\usepackage{multirow}
\usepackage{hyperref}
\usepackage{cite}
\usepackage{url}

\usepackage{xcolor}

\begin{document}
\mainmatter              
\title{Fuzzy Rank-based Late Fusion Technique for Cytology image Segmentation}
\titlerunning{Hamiltonian Mechanics}  
%
\author{Soumyajyoti Dey\inst{1} \and Sukanta Chakraborty \inst{2} \and
Utso Guha Roy\inst{2} \and Nibaran Das\inst{1}}
\authorrunning{Dey et al.} 
%
%
\institute{Jadavpur University, Kolkata, India
\and
Theism Medical Diagnostics Centre, Kolkata, West Bengal, India}

\maketitle              

\begin{abstract}
Cytology image segmentation is quite challenging due to its complex cellular structure and multiple overlapping regions. On the other hand, for supervised machine learning techniques, we need a large amount of annotated data, which is costly. In recent years, late fusion techniques have given some promising performances in the field of image classification. In this paper, we have explored a fuzzy-based late fusion techniques  for cytology image segmentation. This fusion rule  integrates three traditional semantic segmentation models UNet, SegNet, and PSPNet.  The technique is applied on two cytology image datasets, i.e.,  cervical cytology(HErlev) and breast cytology(JUCYT-v1) image datasets.  We have achieved maximum MeanIoU score $84.27\%$ and $83.79\%$ on the HErlev dataset and JUCYT-v1 dataset after the proposed late fusion technique, respectively which are better than that of the traditional fusion rules such as average probability, geometric mean, Borda Count, etc.
The codes of the proposed model are available on \href{https://github.com/DVLP-CMATERJU/cytology_seg}{GitHub}.
\end{abstract}


\section{Introduction}

Cancer is the most common among the deadly diseases\cite{chakraborty2012difficulties} in the twenty-first century. The  rate of cancer is increasing day by day in an alarming ratio. According to 2020 report of World Health Organization\cite{l1}, it is the second leading cause of mortality, and around 10 million people died worldwide from cancer. 
Also, it is predicted that the number of cancer cases in India will increase to 2.08 million\cite{sathishkumar2023cancer} by 2040.
According to the American Cancer Society, 1.9 million new cases will be diagnosed in 2021, and around six lakh patients will die in the USA. In India, most of the cases of cancer have been reported on Breast, Lung, and Cervix. Generally, When patients see lumps on any part of his or her body for a prolonged period, they come to the doctor, and if necessary doctors suggests for some biopsy test. Among different types of biopsy techniques, FNAC (Fine Needle Aspiration Cytology)\cite{ciatto2007accuracy} is a low-cost biopsy technique. In FNAC test,  professional cyto-pathologist analyzes the cytology slides  under microscope or check the cytology images to make a decision (cancer or non-cancerous). 

Over the last few decades, researchers have been tried to develop many CAD (Computer Aided Diagnosis) based systems \cite{saha2016computer, mitra2021cytology, das2023cervical} for the diagnosis of cancer by analyzing cytology images. On cytology images, it consists of many important portions like nuclei, cytoplasm, red blood cells, etc. Also, there are many structural dissimilarities present in cellular objects mainly in the cancerous samples. So, the extraction of the important objects from the entire cytology image is the most challenging task. So, segmentation is the first priority for any CAD-based system. Previously many supervised and unsupervised segmentation methods\cite{chen2021unsupervised, bamford2001method, mohammdian2022optimization, benazzouz2022modified } were applied for cytology image segmentation. Some traditional image segmentation techniques like clustering-based\cite{huang2020nucleus}, filtering-based\cite{win2018comparative}, region-based\cite{zhang2017graph}, etc. have been applied previously for cytology image segmentation. Nowadays, deep learning-based semantic segmentation models are giving promising results on the medical image domain\cite{jiang2022deep}. In that case, we need a large amount of annotated data, but it is much more costly for medical image analysis tasks. However, due to a small amount of annotated data, the semantic segmentation algorithm is not giving satisfactory performance. Sometimes, to improve the segmentation performance, some fusion techniques are used. Previously, the fusion techniques or ensemble rules \cite{dey2023gc} were used mainly for cytology image classification tasks. In this work, we have proposed a fuzzy-based late fusion rule \cite{deb2023breast} to improve the semantic segmentation performances of cytology images. Late fusion(ensemble rule) is a type of rule by which the scores from multiple classifiers are combined\cite{safont2020vector}. Some works\cite{kuiry2020edc3} on late fusion-based segmentation models have already been applied to different natural scene image datasets like CamVid, ICCV09, etc. But in the digital pathology domain mainly in the cytology domain the fuzzy-based fusion of segmentation models is addressed for the first time. However, the fusion of base classifiers is a challenging task, as there is no guarantee that the combinations of base models will boost the segmentation or classification performances than the base classifier. So, as the combination of decisions, some mathematical or statistical models like Dempster–Shafer\cite{nachappa2020flood}, copula\cite{dey2023gc}, average probability\cite{tasci2021voting}, majority voting etc. are introduced.  Recently, different fuzzy-based techniques\cite{basheer2021fesd} have been used for ensemble tasks, to boost the classification performance. In this work, we have used fuzzy rank-based voting technique as a fusion rule for the segmentation of cervical and breast cytology images.

\section{Previous Work}

Over the last few decades, researchers have been trying to develop many automated systems for diagnosing carcinoma by analyzing cytology images. There are lots of works already reported in the cytology image classification domain. However, due to more annotation costs in the supervised learning domain, cytology image segmentation is quite challenging. Previously, some unsupervised techniques were applied for digital pathology image segmentation. Filipczuk et al.\cite{filipczuk2011automatic} proposed an unsupervised technique for breast cytology image segmentation. First, the histogram equalization method is applied on RGB cytology image for contrast enhancement. After that adaptive thresholding method is applied to discard the irrelevant part. Finally, the Gaussian mixture clustering technique to get the final segmentation mask. Along with traditional machine learning techniques, some deep learning-based unsupervised approaches are explored in the digital pathology domain. Zhao et al. \cite{zhao2022lfanet} proposed a cervical cell segmentation technique by extracting different abundant features using a lightweight attention network. 

Recently, some fusion techniques have been adopted in the domain of classification and segmentation. Sen et al.\cite{sen2022ensemble} proposed an ensemble framework for cervical cytology image segmentation by unsupervised approaches. They have explored three clustering techniques, like K-means, K-means++ and Mean Shift as a base segmentation model to find the initial cluster centre. Finally, the decision is combined by Fuzzy C-means clustering algorithm. The proposed technique is evaluated on two  publicly available cervical cytology datasets namely  HervLeV and SIPaKMeD. They have achieved jaccard index score $88.4\%$ on HErlev pap smear dataset.

The major contributions of this work are as follows:

\begin{itemize}
    \item Fuzzy-based late fusion rule for semantic segmentation, by using base semantic segmentation models like UNet, SegNet and PsPNet.

    \item Fusion rule-based segmentation technique is proposed in the domain of breast and cervical cytology.
    
\end{itemize}

\section{Dataset Description:}

In this study, we have used two cytology image database: a) JUCYT-v1\footnote{$https://github.com/DVLP-CMATERJU/JUCYT_V1$}, b) HErlev Pap Smear\cite{basak2021cervical}.

\begin{figure}[h]
    \centering
    \includegraphics[width=\textwidth]{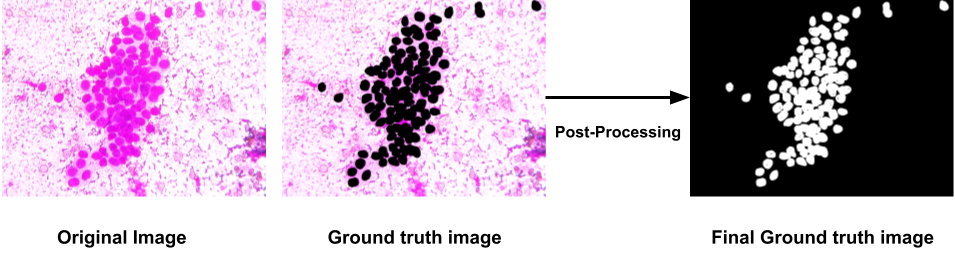}
    \caption{Ground truth preparation of JUCYT-v1 dataset}
    \label{fig:fig1}
    
\end{figure}

JUCYT-v1 is a database of breast cytology images, which consists of 62 cytology images(39 non-cancerous, 23 cancerous). The cytology images by FNAC test are collected from the medical diagnosis center named "Theism Diagnosis Centre, Kolkata, India" in the presence of a professional practitioner. The slides are viewed under the trinocular microscope, and the ROIs of 40x magnification are captured by a 5 MP CMOS camera. For supervised semantic segmentation, we need pixel-level annotations. 
So these RGB cytology images are manually labeled by the photo-shop software. After layer-wise marking of each cell of cytology images, some post-processing techniques are applied for fine-tuning the ground truths. Finally, in the ground truth mask the white pixels and black pixels represent the informative portions and background respectively.  The flow diagram of the final segmentation mask(ground truth) preparation is described in Fig.  \ref{fig:fig1}. 
For the experiment, it is split into 4:1 ratio, i.e. 50 images for training and 12 images for testing. Some samples of the JUCYT-v1 dataset along with corresponding ground truths are mentioned in Fig. \ref{fig:fig3}. 

\begin{figure}[h]
    \centering
    \includegraphics[width=\textwidth]{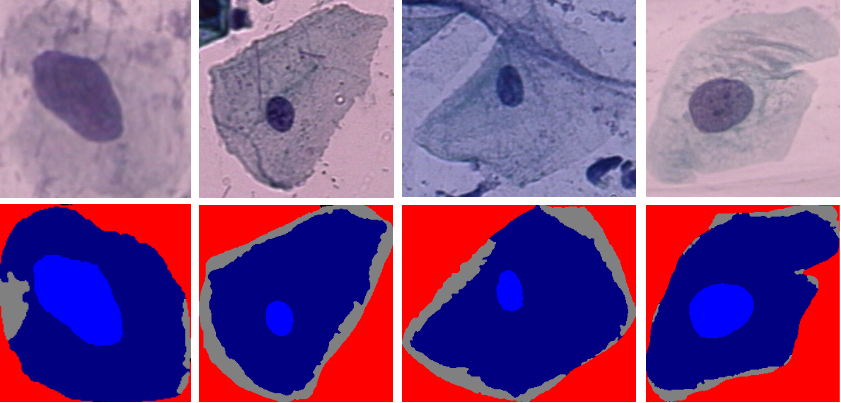}
    \caption{Cervical Cytology Data(HErlev).Top row indicated the original images, Bottom row indicates the corresponding ground truth images }
    \label{fig:fig2}
    
\end{figure}

\begin{figure}[h]
    \centering
    \includegraphics[width=\textwidth]{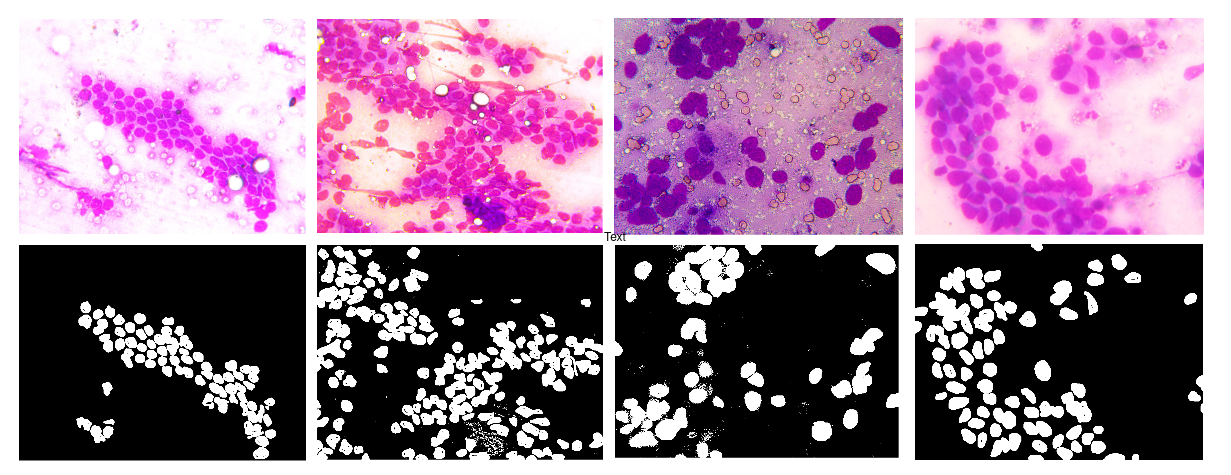}
    \caption{Breast Cytology Data(JUCYT-v1).Top row indicates the original images, Bottom row indicates the corresponding ground truth images }
    \label{fig:fig3}
    
\end{figure}

HErlev is a publicly available cervical cytology image database, which consists of 917 pap smear images.  The  dataset is divided into train, test, and validation sets in the ratio of 3:1:1 (i.e. 550 train, 182 test, and 185 validation set). The dataset consists of five different labels from the perspective of sementic segmentation. The ground truth of each image is labeled with five different colors, and these are indicative of five different portions of the Pap smear image. The red color indicates the background portions, dark blue indicates the cytoplasm portions, light blue indicates the nuclei portions, etc. Some samples of the HErlev pap smear dataset are mentioned on Fig. \ref{fig:fig2}.

\section{Proposed Methodology} 

\begin{figure}[h]
    \centering
    \includegraphics[width=\textwidth]{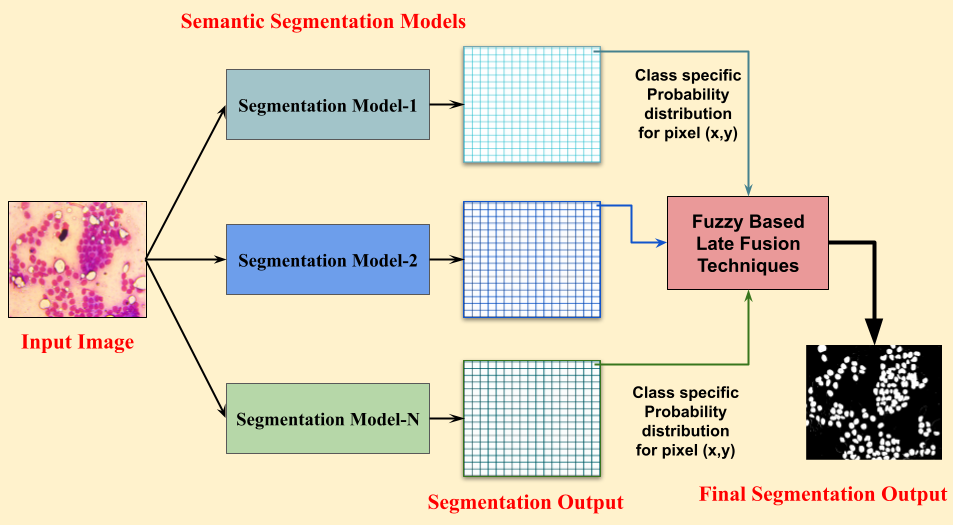}
    \caption{Outline of proposed Fuzzy based Late fusion rule for cytology image segmentation}
    \label{fig:fig4}
    
\end{figure}

In this paper, we have proposed a fuzzy-based late fusion rule for the semantic segmentation of digital cytology images. The proposed methodology consists of two phases:
a) cytology image segmentation by supervised semantic segmentation models; b) fuzzy-based late fusion of semantic segmentation models to boost the segmentation performances compared to the base models. The flow chart of the proposed methodology is described in Fig. \ref{fig:fig4}.

\subsection{Semantic Segmentation by Traditional Deep Learning Models }

Semantic Segmentation is the technique where every pixel of the image is classified by some machine learning or deep learning models. It has very trending methodology in the field of deep learning for the last few years. In the proposed study we have used traditional supervised models like UNet\cite{azad2022medical}, SegNet\cite{badrinarayanan2017segnet}, and PSPNet\cite{zhou2019fusion} as base semantic segmentation models.

Let $I_(224\times224)$ be the cytology image and $M_1, M_2, M_3$ be the UNet, SegNet, and PSPNet models, respectively.
Therefore, for each pixel $(x,y)$ of the image $I_(224\times224)$, we have evaluated the scores $\{\ y^j_1,y^j_2,.....,y^j_C\ \}$ with respect to each model $M_j$ for all j and $C$ be the no of classes(for HErlev $C$=5 and for JUCYT-v1 $C$=2).

$\{\ y^j_1,y^j_2,.....,y^j_C\ \} \leftarrow M_j(I_(224\times224(x,y)))$

The pixel $(x,y)$ is predicted by $argmax(\{\ y^j_1,y^j_2,.....,y^j_C\ \})$

\textbf{Training Process of Segmentation:} The cytology images are resized at the dimension of 224$\times$224 pixels. At the time of CNN (Convolution Neural Network) training, some hyperparameters like batch size, number of epochs, and the learning rate are set to 16, 1000, and 0.0001 respectively. Here ADAM optimizer is used for all the networks. The training loss is calculated by the negative log-likelihood estimation method. The best-trained models are saved when the validation loss is minimal. 

After evaluating the best train models, the class-specific probability distribution values for each pixel of the test set image are now evaluated by the SOFTMAX function. 

where, $SOFTMAX (y) = \frac{e^y}{\sum e^y}$

These probability distribution values are used for next-level fusion techniques.

\subsection{Fuzzy based Late Fusion Rule:}
 Let $M_j$ be the supervised segmentation model, where $j=1,2,3,..N$ and $N$ be the total number of segmentation models.
Assume that, $\{\ P^j_1,P^j_2,.....,P^j_C\ \}$ be the set of class specific probability distribution values of the pixel co-ordinate $(x,y)$ for the test image $I$ with respect to segmentation model $M_j$ and $C$ be the total number of class, where
$\sum\limits_{i=1}^C P_i^j = 1$, $j=1,2,..,N$.

\subsubsection{Fuzzy Rank based Voting Rule:}

Fuzzy rank-based voting is a fusion rule where the confidence scores of base learners are fused by membership functions.

The two sets of fuzzy rank for segmentation model $M_j$ be  $\{\ r_1^{j_1}, r_2^{j_1},...., r_C^{j_1} \}\ $ and $\{\ r_1^{j_2}, r_2^{j_2},...., r_C^{j_2} \}\ $, which are generated by two non-linear functions such that, 
$r_k^{j_1} = 1- tanh(\frac{({P_k^j}-1)^2}{2})$ and $r_k^{j_2} = 1- exp(\frac{-({P_k^j}-1)^2}{2})$, for each class $k$. Now assume the set of fused rank scores be $\{\  {rs}_1^j, {rs}_2^j,....., {rs}_C^j \}\ $ for the model $M_j$ where, ${rs}_k^j = r_k^{j_1} \times r_k^{j_2}$, for each class $k$.
 Let, $ \{\ {fs}_1, {fs}_2,....., {fs}_C  \}\ $ be the set of fused score where ${fs}_k = \sum\limits_{j=1}^N {rs}_k^j$ for each class $k$. Finally the predicted class of the pixel $(x,y)$ be 
 $\textbf{min} \{\ {fs}_1, {fs}_2,....., {fs}_C  \}\ $.

 This operation is executed for all the pixels of the text image $I_{224\times224}$

\section{Results and Discussion}

\begin{figure}[h]
    \centering
    \includegraphics[width=\textwidth]{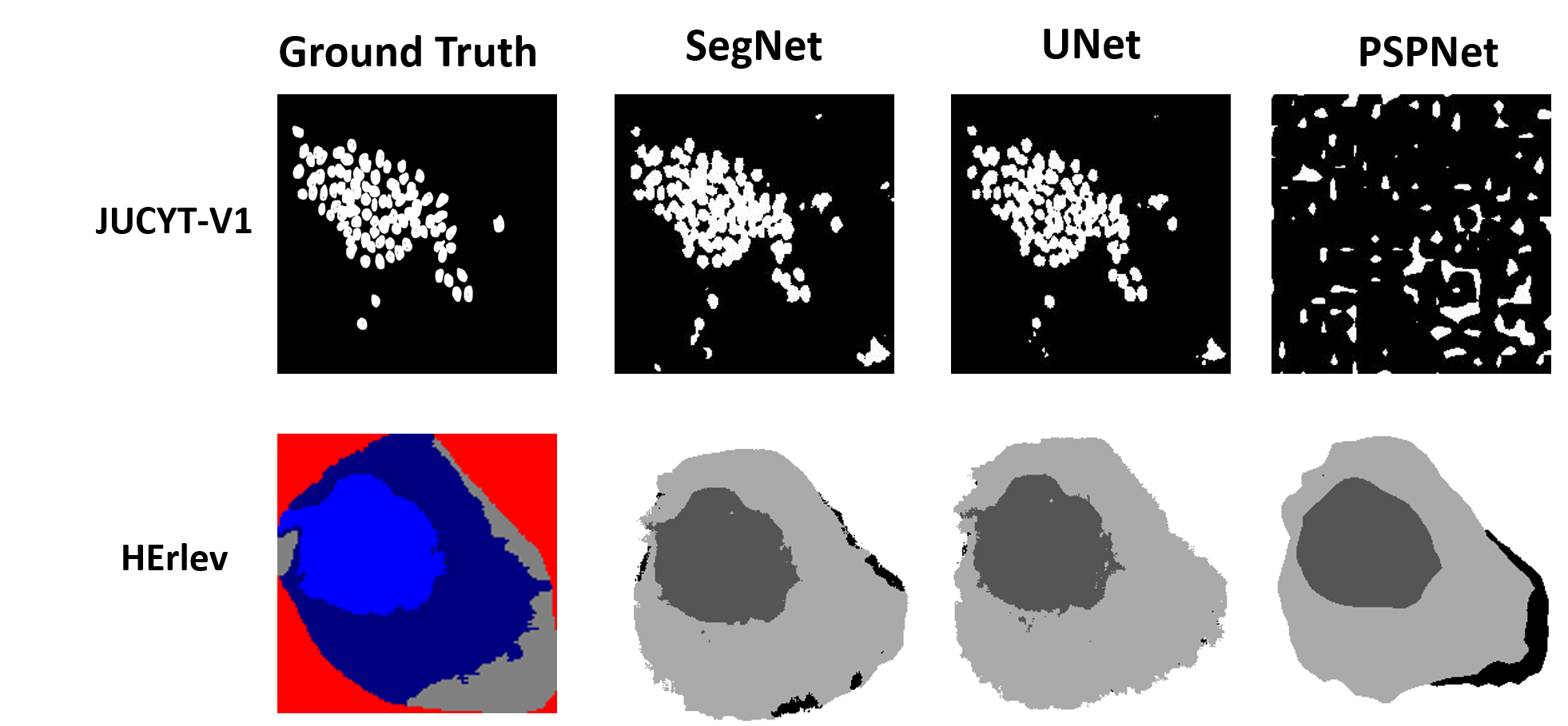}
    \caption{ Output segmentation masks evaluating by base semantic Segmentation models}
    \label{fig:figbase}
    
\end{figure}

The main target of this proposed model is to perform semantic segmentation based on the joint performances of the combination of base semantic segmentation models. In table \ref{tab:tab1}, we have described the segmentation performance of base models for the two cytology databases. For HErlev dataset, the PSPNet is performing better than SegNet and UNet models. But for the JUCYT-v1 dataset, UNet is performing best as a base classifier. After the fuzzy rank based fusion, we got 84.27 \% meanIoU in the combination of UNet and PSPNet on HErlev dataset. It has improved 0.93\% segmentation performance than the top base classifier mean IoU score. But for the JUCYT-v1 dataset, we have achieved 83.79\% mean IoU score in the combination of UNet and SegNet model. It has increased the segmentation performance by approximately 5.9\% after the proposed fusion technique on JUCYT-v1 dataset. Due to less amount of data in JUCYT-v1, we have reported the result on the validation set. Also, we have only reported the joint performances of the SegNet and UNet for JUCUT-v1 dataset. Also, it is shown that at the base level, PSPNet is performing worst and it is not learning the proper structures of the breast cytology dataset. The segmentation masks generated by base semantic segmentation models are described on Fig. \ref{fig:figbase}.

\subsection{Comparative study}

Also, we have made a comparative study with traditional fusion rules by the performance of semantic segmentation on the cytology dataset. The descriptions of traditional late fusion techniques are as follows:

\subsubsection{Arithmetic Average rule:}

Arithmetic average rule is a voting rule, where the mean value of class wise probability distribution values of different models are evaluated. 

$ P^k_{avg}=\frac{1}{j}\sum\limits_{j=1}^N(P_k^j)$ , where $ P^k_{avg}$ be the average probability of the class $k$. 
Therefore the predicted class of the pixel $(x,y)$ be  $ \textbf{argmax}\{\ P^1_{avg}, P^2_{avg},.......,P^C_{avg}  \}\ . $

\subsubsection{Geometric Average rule:}
In geometric average rule, the product value of class wise probability distribution values are evaluated.

$ P^k_{prod}=\frac{1}{j}\Pi_{j=1}^N(P_k^j)$ , where $ P^k_{prod}$ be the product of probabilities of the class $k$. 
Therefore the predicted class of the pixel $(x,y)$ be  $ \textbf{argmax}\{\ P^1_{prod}, P^2_{prod},.......,P^C_{prod}  \}\ . $ 

\subsubsection{Median value rule:}

In median value rule, the median value of class wise probability distribution values are evaluated.

$ P^k_{med}=\textbf{Median}\{\ P_k^1,  P_k^2,  P_k^3,....,  P_k^N\}\ $ , where $ P^k_{med}$ be the median value of probabilities of the class $k$. 
Therefore the predicted class of the pixel $(x,y)$ be \\ $ \textbf{argmax}\{\ P^1_{med}, P^2_{med},.......,P^C_{med}  \}\ . $ 

\subsubsection{Max Rule}

In max voting rule, the maximum value of class specific probability distribution values are computed.
$ P^k_{max}=\textbf{max}\{\ P_k^1,  P_k^2,  P_k^3,....,  P_k^N\}\ $ , where $ P^k_{max}$ be the maximum value of probabilities of the class $k$. Therefore the predicted class of the pixel $(x,y)$ be $ \textbf{argmax}\{\ P^1_{max}, P^2_{max},.......,P^C_{max}  \}\ . $

\subsubsection{Min Rule}

In max voting rule, the minimum value of class-specific probability distribution values are computed.
$ P^k_{min}=\textbf{min}\{\ P_k^1,  P_k^2,  P_k^3,....,  P_k^N\}\ $ , where $ P^k_{min}$ be the minimum value of probabilities of the class $k$. Therefore the predicted class of the pixel $(x,y)$ be $ \textbf{argmax}\{\ P^1_{min}, P^2_{min},.......,P^C_{min}  \}\ . $ 

\subsubsection{Borda Count(BC) rule:} 

In BC rule, first the class specific probability distribution values of the pixel $(x,y)$ with respect to the model $M_j$, are arranged in descending order.\\ 
$ \{\ r^j_1, r^j_2,...., r^j_C \}\ \leftarrow \textbf{Descending}_{C}\{\ P^j_1,P^j_2,.....,P^j_C\ \}$ where, $r^j_k$ be the rank of the class $k$ with respect to classifier $j$ after arranging the distribution values in descending order. Finally the voting score ($v_k$) is calculated for the class $k$ by $ v_k = \sum\limits_{j=1}^N(C-r^j_k)$, for each class $k$. Therefore the predicted class for the pixel $(x,y)$ be $\textbf{max} \{\ v_1, v_2,...., v_C \}\ $.

By combining various base segmentation models, Table \ref{tab:t2} and \ref{tab:t3} describe the segmentation performances of various late fusion rules. 

\begin{figure}[h]
    \centering
    \includegraphics[width=\textwidth]{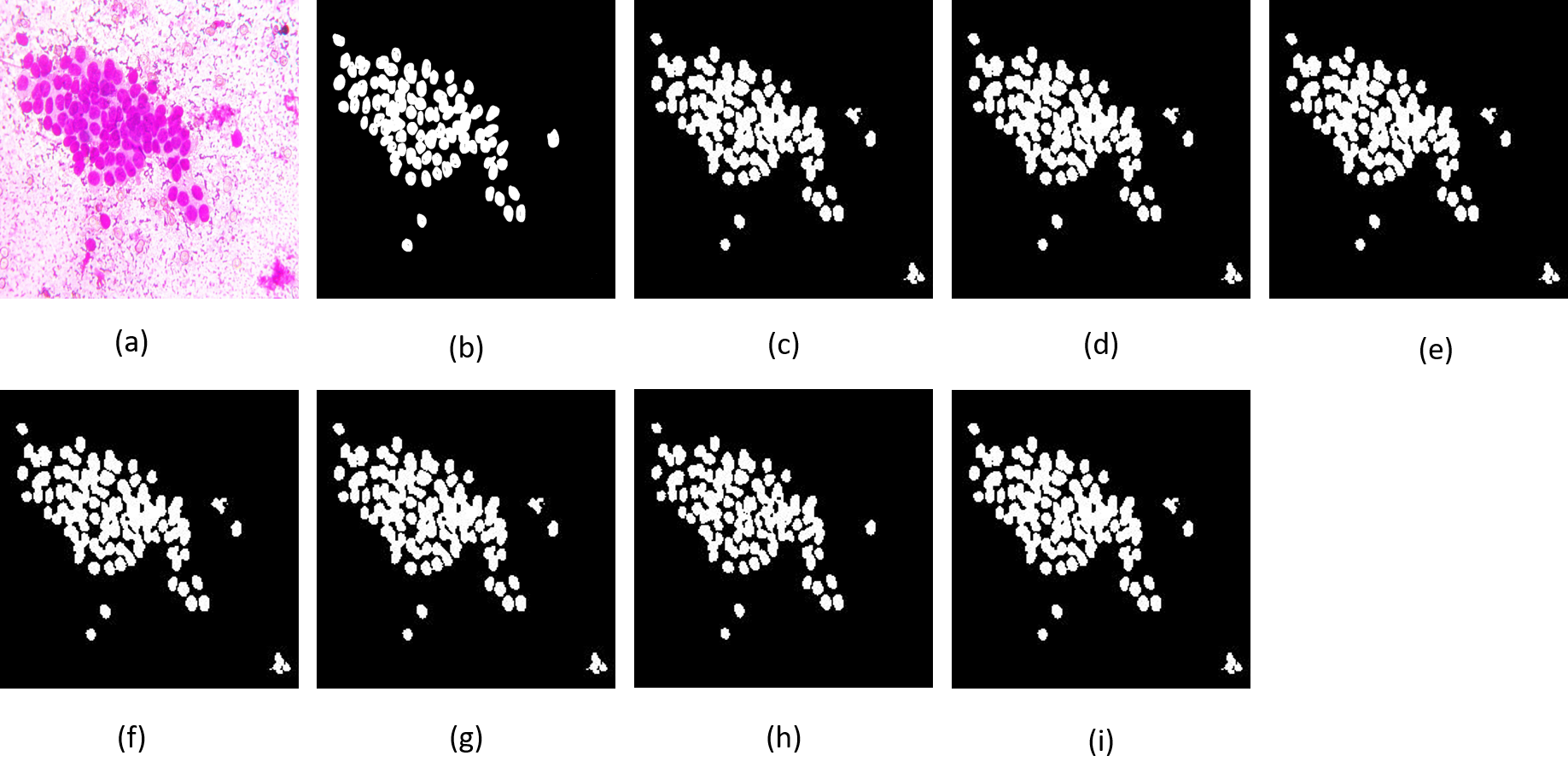}
    \caption{Segmentation masks of different fusion  techniques(Late fusion of UNet and SegNet) of JUCYT-v1 dataset (a) Original Image, (b) Ground Truth, (c) Arithmetic Average, (d) Geometric Average, (e) Median Rule, (f) Max Rule, (g) Min Rule, (h) Borda Count Rule, (i) Fuzzy Rank based Voting rule }
    \label{fig:fig5}
    
\end{figure}

\begin{figure}[h]
    \centering
    \includegraphics[width=\textwidth]{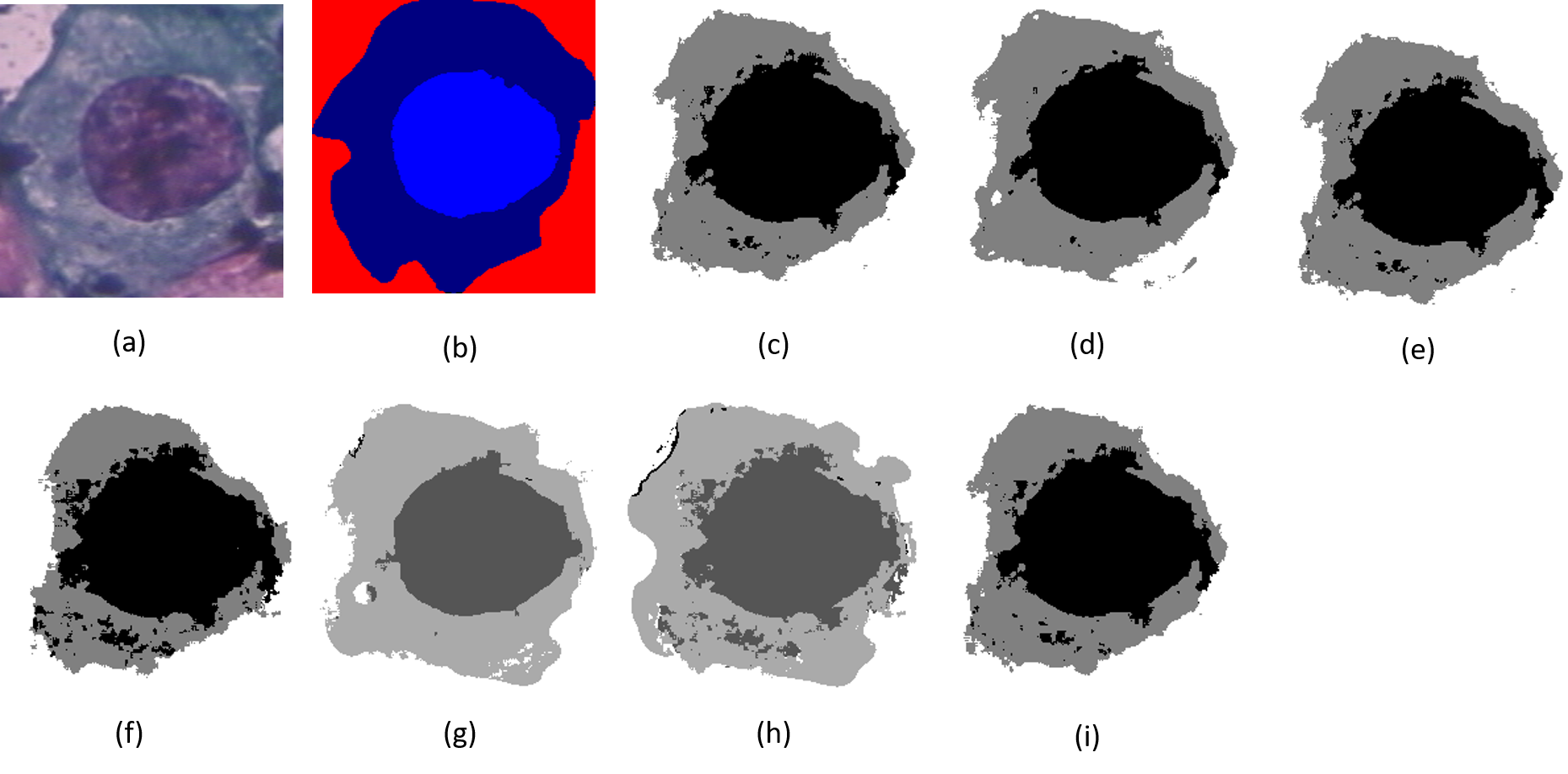}
    \caption{Segmentation masks of different fusion techniques(Late fusion of UNet and SegNet) of HErlev Dataset(a) Original Image, (b) Ground Truth, (c) Arithmetic Average, (d) Geometric Average, (e) Median Rule, (f) Max Rule, (g) Min Rule, (h) Borda Count Rule, (i) Fuzzy Rank based Voting rule }
    \label{fig:fig6}
    
\end{figure}

Also, the segmentation outputs of different late fusion rules are described in Fig. \ref{fig:fig5} and Fig. \ref{fig:fig6}.

\begin{table}[]
\caption{Segmentation Performance of Base Semantic Segmentation Model}
\label{tab:tab1}
\resizebox{0.5\columnwidth}{!}{%
\begin{tabular}{|c|c|c|}
\hline
\textbf{Dataset} & \textbf{Model} & \textbf{Mean IoU} \\ \hline
\multirow{3}{*}{HErlev} & U-Net & 81.58 \\ \cline{2-3} 
 & Seg-Net & 71.77 \\ \cline{2-3} 
 & PSP-Net & 83.34 \\ \hline
\multirow{3}{*}{JUCYT-v1} & U-Net & 77.88 \\ \cline{2-3} 
 & Seg-Net & 66.3 \\ \cline{2-3} 
 & PSP-Net & 56.8 \\ \hline
\end{tabular}%
}
\end{table}

\begin{table}[]
\caption{Different Fusion Rule based performances on HErlev  dataset(U= U-Net, S= Seg-Net, P=PSP-Net)}
\label{tab:t2}
\resizebox{\columnwidth}{!}{%
\begin{tabular}{|c|c|c|c|c|c|c|c|}
\hline
\textbf{MODEL} & \textbf{\begin{tabular}[c]{@{}c@{}}Average\\  Probability\end{tabular}} & \textbf{\begin{tabular}[c]{@{}c@{}}Geometric \\ Mean\end{tabular}} & \textbf{Median} & \textbf{Maxrule} & \textbf{Minrule} & \textbf{BC-Rule} & \textbf{\begin{tabular}[c]{@{}c@{}}Fuzzy Rank\\ based Voting\end{tabular}} \\ \hline
\textbf{U+S+P} & 83.42 & 83.41 & 82.98 & 82.62 & 82.92 & 82.90 & 83.38 \\ \hline
\textbf{U+S} & 75.91 & 77.77 & 75.91 & 72.25 & 79.52 & 75.20 & 75.19 \\ \hline
\textbf{U+P} & 84.11 & 84.01 & 84.11 & 83.39 & 84.05 & 83.06 & \textbf{84.27} \\ \hline
\textbf{P+S} & 82.68 & 82.56 & 82.68 & 82.62 & 82.19 & 79.24 & 82.64 \\ \hline
\end{tabular}%
}
\end{table}

\begin{table}[]
\caption{Different Fusion Rule based performances on JUCYT-v1  dataset(U= U-Net, S= Seg-Net), P= PSP-Net}
\label{tab:t3}
\resizebox{\textwidth}{!}{%
\begin{tabular}{|c|c|c|c|c|c|c|c|}
\hline
\textbf{MODEL} & \textbf{\begin{tabular}[c]{@{}c@{}}Average\\ Probability\end{tabular}} & \textbf{\begin{tabular}[c]{@{}c@{}}Geometric\\ Mean\end{tabular}} & \textbf{Median} & \textbf{Maxrule} & \textbf{Minrule} & \textbf{BC-Rule} & \textbf{\begin{tabular}[c]{@{}c@{}}Fuzzy Rank\\ based Voting\end{tabular}} \\ \hline
U+S+P          & 70.84                                                                  & 70.24                                                             & 70.82           & 68.19            & 68.19            & 70.82            & 71.25                                                                      \\ \hline
U+S            & 74.25                                                                  & 74.25                                                             & 74.25           & 79.25            & 74.25            & 66.56            & \textbf{83.79}                                                             \\ \hline
U+P            & 66.42                                                                  & 66.42                                                             & 66.42           & 66.42            & 66.42            & 56.8             & 66.42                                                                      \\ \hline
P+S            & 61.69                                                                  & 61.69                                                             & 61.69           & 61.69            & 61.69            & 53.69            & 61.69                                                                      \\ \hline
\end{tabular}%
}
\end{table}

\section{Conclusion}

In this paper, we have explored different late fusion rules in the domain of semantic segmentation. On the cervical cytology dataset, we have achieved 84.27\% segmentation performance. Also, in breast cytology domain, it has improved remarkable performances in the fuzzy based fusion of UNet and SegNet models. In future, we will explore other fuzzy-based rules, to improve segmentation performance. Also, due to the limited amount of cytology data, we will explore other semi-supervised or weakly supervised learning techniques for cytology image segmentation. In future, we will try to introduce some transformer based segmentation models like swin transformer, vit, etc. as a base level classifier to improve the symantic segmentation model.

\section{Acknowledgement}

The authors are thankful to CMATER Lab, CSE Department, Jadavpur University for providing the infrastructural support during the experiment. Also, thankful to Theism Medical Diagnostics Centre, for providing the breast cytology samples. This work is financially supported by SERB (DST), Govt. of India (Ref. No.: EEQ/2018/000963). 

\bibliographystyle{plain}
\bibliography{ref.bib}

\end{document}